\pdfoutput=1

\documentclass[11pt]{article}

\usepackage{ACL2023}

\usepackage{times}
\usepackage{latexsym}
\usepackage{amsmath,amssymb,amsfonts}
\usepackage{todonotes}
\usepackage{bm}
\usepackage[ruled]{algorithm2e}
\usepackage[T1]{fontenc}

\usepackage[utf8]{inputenc}

\usepackage{microtype}

\usepackage{inconsolata}

\title{Exploring Multilingual Text Data Distillation}

\author{
    Shivam Sahni \\
  \texttt{ssahni@ucsd.edu} \\\And
    Harsh Patel \\
  \texttt{h1patel@ucsd.edu} \\
  }

\begin{document}
\maketitle
\begin{abstract}

With the rise of deep learning, large datasets and complex models have become common, requiring significant computing power. To address this, data distillation has emerged as a technique to quickly train models with lower memory and time requirements. However, data distillation on text-based datasets hasn't been explored much because of the challenges rising due to its discrete nature. Additionally, existing dataset distillation methods often struggle to generalize to new architectures. In the paper, we propose several data distillation techniques for multilingual text classification datasets using language-model-based learning methods. We conduct experiments to analyze their performance in terms of classification strength, and cross-architecture generalization.  Furthermore, we investigate the language-specific fairness of the data summaries generated by these methods. Our approach builds upon existing techniques, enhancing cross-architecture generalization in the text data distillation domain. Our code is available at \url{https://github.com/Harshp1802/text-dataset-distillation}
\end{abstract}

\section{Introduction}

The widespread use of deep learning has led to the creation of large datasets and the development of complex models that require massive amounts of parameters. However, due to limited computing power, there is a significant need for alternatives that can improve these models' memory and time efficiency. In the past, methods have relied on knowledge distillation to reduce the size of these models while maintaining their performance \cite{sanh2020distilbert}. Recently, data distillation has emerged as another viable option. Instead of reducing the model's size, data distillation aims to create a compact summary of the original dataset that can act as a proxy for training a model from scratch. This approach can lead to faster model training, facilitating faster experimentation, low-resource training, and lower energy consumption.

In this paper, we address four key challenges that hinder the widespread application of dataset distillation methods.

First, we explore the application of dataset distillation to text-based datasets. While dataset distillation has demonstrated remarkable performance on image-based datasets \cite{wang2020dataset, cazenavette2022distillation, zhao2021dataset}, its effectiveness on text-based datasets remains limited. The discrete nature of text data poses a challenge in directly applying gradient-based methods for distillation  \cite{sachdeva2023data}. To overcome this, we use a common technique in NLP-based methods i.e., embedding the text into a continuous space to make it amenable to text data distillation \cite{Sucholutsky_2021}. In this work, we use text classification datasets and pre-trained language-model-based classification models for performing dataset distillation.

Second, we extend dataset distillation to multilingual data, enabling us to perform language-specific analyses on the distilled datasets. This can be helpful to evaluate whether the distilled data suffers from the fairness concerns commonly involved in biased datasets \cite{De_Arteaga_2019}. We analyze the distribution of samples across different languages in the distilled data and evaluate whether the model trained on the distilled data performs similar to the original model for all languages.

Third, we address the challenge of cross-architecture generalization. Synthetic datasets generated through distillation are often optimized for specific network architectures, limiting their ability to generalize to different architectures \cite{cazenavette2023generalizing}. We propose different approaches to text data distillation that potentially enhance the generalization capability of the synthetic datasets, enabling them to perform well across various network architectures.

Lastly, we conduct an in-depth analysis of the interpretability of the dataset distillation methods proposed in this study. While distillation techniques work impressively at compressing information from large datasets, it is essential to assess the interpretability of the generated summaries. In the case of text-based distillation, the distilled data is generated in the embedding space, which allows for interpretation by identifying the nearest sentences that correspond to the distilled embeddings. We demonstrate the interpretability of the proposed methods by showcasing how this process can be performed for each of the proposed techniques.

By addressing these key directions, we expand the applicability and effectiveness of text dataset distillation methods, particularly in the context of fairness, interpretability, and  cross-architecture generalization.

\subsection{Problem Statement}
Dataset Distillation is the task of synthesizing a smaller data summary that can distill the most pertinent information from a given large dataset. Using such a condensed data summary in practice has several advantages, such as accelerating model training, being sustainable, and using less storage.

Following \cite{sachdeva2023data}, let $\mathcal{D} \triangleq\left\{\left(x_i, y_i\right)\right\}_{i=1}^{|\mathcal{D}|}$ be a given dataset which needs to be distilled, where $x_i$ are the set of input features, and $y_i$ is the desired label for $x_i$. Given a data budget $n$, data distillation aims to synthesize a data summary $\mathcal{D}_{\mathrm{syn}} \triangleq\left\{\left(\tilde{x}_i, \tilde{y}_i\right)\right\}_{i=1}^n$ such that $n \ll|\mathcal{D}|$. Given a learning algorithm $\Phi$, let $\theta^{\mathcal{D}}, \theta^{\mathcal{D}_{\text {syn }}}$ represent the optimal set of parameters for $\Phi$ estimated on $\mathcal{D}$ and $\mathcal{D}_{\mathrm{syn}}$; data distillation can be defined as optimizing the following:

\begin{equation}
    \begin{split}
        \underset{\mathcal{D}_{\text {syn }, n}}{\arg \min }& (\sup \{|l(\Phi_{\theta^{\mathcal{D}}}(x), y)- \\ &\quad l(\Phi_{\theta^{\mathcal{D}_{\mathrm{syn}}}}(x), y)|\}_{\substack{x \sim \mathcal{X} \\ y \sim \mathcal{Y}}})
    \end{split}
\end{equation}


\section{Related Work}
\label{sec:relatedwork}
Wang et al. \cite{wang2020dataset} are the first to propose a method to distill a large dataset into a smaller one. Their main idea was to synthesize a small number of images that need not come from the original data distribution, but when given to a learning algorithm as training data, it approximates the model trained on the original dataset. They were able to show that it is possible to compress 60,000 MNIST training images into just 10 synthetic distilled images and achieve close to original performance with just a few gradient descent steps, given a fixed network initialization.

\begin{algorithm}
    \caption{Text Data Distillation}
    \label{alg:tdd}
    \KwData{$p(\theta_0)$: distribution of initial weights; $M$: the number of distilled data; $\alpha$: step size; $n$: batch size; $T$: number of optimization iterations; $\tilde{y_0}$: initial value for $\tilde{y}$; $\tilde{\eta}_0$: initial value for $\tilde{\eta}$; $s$: sentence length; $d$: embedding size}
    Initialize distilled data \\
    $\tilde{\mathbf{x}} = \{\tilde{x}_i\}_{i=1}^M$ randomly of size $s \times d$ \\
    $\tilde{\mathbf{y}} = \{\tilde{y}_i\}_{i=1}^M \gets \tilde{y_0}$ \\
    $\tilde{\eta} \gets \tilde{\eta}_0$\\
    \For{each training step $t = 1$ to $T$}{
        Get a mini-batch of real training data $\mathbf{x}_t = \{x_{t, j}, y_{t, j}\}_{j=1}^n$\\
        Pad (or truncate) each sentence in the mini-batch $\left(\mathbf{x^P}_t, \mathbf{y}_t\right)=\left\{\operatorname{Pad}\left(x_{t, j}, \text { len }=s\right), y_{t, j}\right\}_{j=1}^n$ \\
        Embed each sentence in the mini-batch $n$ $\left(\mathbf{x}^*_t, \mathbf{y}_t\right)=\left\{\operatorname{Embed}\left(x_{t, j}^p, \operatorname{dim}=d\right), y_{t, j}\right\}_{j=1}^n$ \\
        One-hot encode the labels $(\mathbf{x^*}_t, \mathbf{y}_t^*)= \{x^*_{t, j},$Encode$(y_{t, j})\}_{j=1}^n$ \\
        Sample a batch of initial weights $\theta_0^{(j)} \sim p\left(\theta_0\right)$ \\
        \For{each sampled $\theta_0^{(j)}$}{
            Compute updated model parameter with GD $\theta_1^{(j)}=\theta_0^{(j)}-\tilde{\eta} \nabla_{\theta_0^{(j)}} \ell\left(\tilde{\mathbf{x}}, \tilde{\mathbf{y}}, \theta_0^{(j)}\right)$ \\
            Evaluate the objective function on real training data: $\mathcal{L}^{(j)}=\ell\left(\mathbf{x^*}_t, \mathbf{y^*}_t, \theta_1^{(j)}\right)$ \\
        }
        Update distilled data \\
        $\tilde{\mathbf{x}} \leftarrow \tilde{\mathbf{x}}-\alpha \nabla_{\tilde{\mathbf{x}}} \sum_j \mathcal{L}^{(j)}$ \\
        $\tilde{\mathbf{y}} \leftarrow \tilde{\mathbf{y}}-\alpha \nabla_{\tilde{\mathbf{y}}} \sum_j \mathcal{L}^{(j)}$ \\
        $\tilde{\eta} \leftarrow \tilde{\eta}-\alpha \nabla_{\tilde{\eta}} \sum_j \mathcal{L}^{(j)}$ \\
    }
    \For{$i = 1 \text{ to } M$}{
        Compute nearest embedding for every distilled word $\tilde{\mathbf{x}}_i^*=\left\{\operatorname{NearestEmbed}\left(\tilde{x}_{i, j}\right)\right\}_{j=1}^s$ \\
        Decode embedding into text $\tilde{\mathbf{z}}_i=\left\{\operatorname{WordFromEmbed}\left(\tilde{x}_{i, j}^*\right)\right\}_{j=1}^s$
    }
    $\tilde{\mathbf{z}}=\left\{\tilde{z}_i\right\}_{i=1}^M$ \\
     \KwResult{distilled data $\tilde{\mathbf{x}}$, distilled labels $\tilde{\mathbf{y}}$, optimized learning rate $\tilde{\eta}$, nearest sentences $\tilde{\mathbf{z}}$}
\end{algorithm}

\cite{Sucholutsky_2021} extend the data distillation algorithm to distill sequential datasets including text by embedding it into a continuous space. They show that text dataset distillation outperforms competing techniques across several datasets. For instance, a model trained over distilled IMDB sentiment classification dataset with just ten instances per class achieves 97.8\% of the original accuracy. They use glove-based word vectors while embedding text into continuous space. This enables the use of the nearest-neighbor method to identify the most similar words from the original dictionary. In this work, we focus on language-model-based embedding spaces for text. Therefore, we introduce different variations of nearest-neighbor methods to assess their interpretability. Algorithm \ref{alg:tdd} outlines the text dataset distillation algorithm \cite{Sucholutsky_2021}. This algorithm serves as the fundamental framework for all of our methods.  

Han et al. \cite{han-etal-2022-towards-fair} further investigate how the data distillation approach impacts group bias with experiments over two language classification tasks. Their findings indicate that conventional data distillation methods tend to maintain the biases present in the original dataset. In our research, we extend this investigation by experimenting with multilingual data, enabling us to perform similar analyses with language-specific fairness in mind.

The field of multilingual natural language processing has gained significant attention in recent years. Some advancements include models such as mBERT \cite{devlin-etal-2019-bert} and XLM-R \cite{conneau2020unsupervised}, which have demonstrated exceptional performance across a wide range of multilingual tasks. These models are pretrained on large-scale multilingual corpora, enabling them to effectively capture linguistic patterns across several languages. These models find application in various domains, including but not limited to machine translation, cross-lingual information retrieval, sentiment analysis, and many others. The success of mBERT and XLM-R motivates us to employ them in dataset distillation to enhance language comprehension.

Recent works in dataset distillation have emphasized cross-architecture generalization as a crucial metric. \cite{NEURIPS2022_07bc722f} introduce a novel dataset factorization technique to enhance the representation capability of datasets. The distilled datasets produced using this approach achieved a remarkable 10\% higher accuracy compared to baseline methods in terms of cross-architecture generalization. More recently, \cite{cazenavette2023generalizing} proposed Generative Latent Distillation (GLaD), which utilizes a deep generative prior to parameterize the synthetic dataset in the intermediate feature space of generative models such as Generative Adversarial Networks (GANs). This prior encourages the learned datasets to exhibit enhanced generalizability to new architectures. However, it is important to note that both of these approaches primarily focus on image-based datasets.

To the best of our knowledge, our work is the first to address cross-architecture generalization for text-based dataset distillation.  

\section{Methodology}
\label{sec:methods}
We will now explore our four distinct approaches, each of which employs a unique methodology for generating the distilled data.

\subsection{VanillaDistill: Vanilla method with mBERT }
\label{sec:vanilladistill}
\begin{figure}[]
\centering
\includegraphics[width=\linewidth]{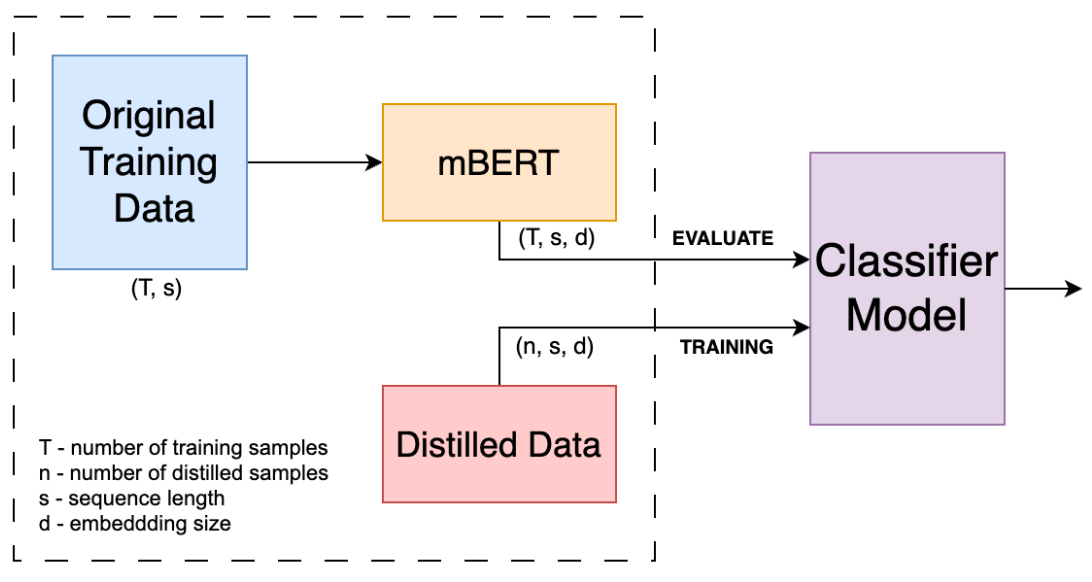}
\caption{VanillaDistill Model Architecture: Here we use mBERT embeddings rather than Glove \cite{Sucholutsky_2021}. The algorithm comprises two steps: 1) Training Step, where a classifier is trained on distilled data, and 2) Evaluation Step, which assesses the trained classifier on real training data to obtain an evaluation loss. This evaluation loss is then used to backpropagate gradients and update the distilled data.} 
\label{fig:VanillaDistill}
\end{figure}
In this method, we employ a basic data distillation pipeline that utilizes a pre-trained language model (specifically, a frozen mBERT) with a classification head. This method differentiates itself from existing text-based data distillation techniques (as discussed in Section \ref{sec:relatedwork}) by leveraging contextual embeddings generated by the language model, as opposed to non-contextual Glove-based word vectors.

Figure \ref{fig:VanillaDistill} illustrates the data distillation pipeline for our method, known as VanillaDistill. Notably, in this approach, the distilled embeddings are learned at the output level of the language model. Considering the interpretability aspect, where we aim to map each learned word embedding to a word in the original dictionary, this approach poses a problem due to the absence of ground truth embeddings for the words in our original dictionary at the output level. This realization serves as a motivation for us to develop more robust and effective methods, which we discuss in the subsequent sections. To estimate the output-level ground truth mBERT word embeddings for the VanillaDistill method, we employ a heuristic where we individually input each word from the training sample into the mBERT model to obtain the corresponding output-level word embedding. These word embeddings can further be used to find the nearest words of our distilled data.

\subsection{SkipLookupDistill: Distillation at mBERT’s input-level}
\label{meth:skiplookup}

\begin{figure*}[]
\centering
\includegraphics[width=0.7\linewidth]{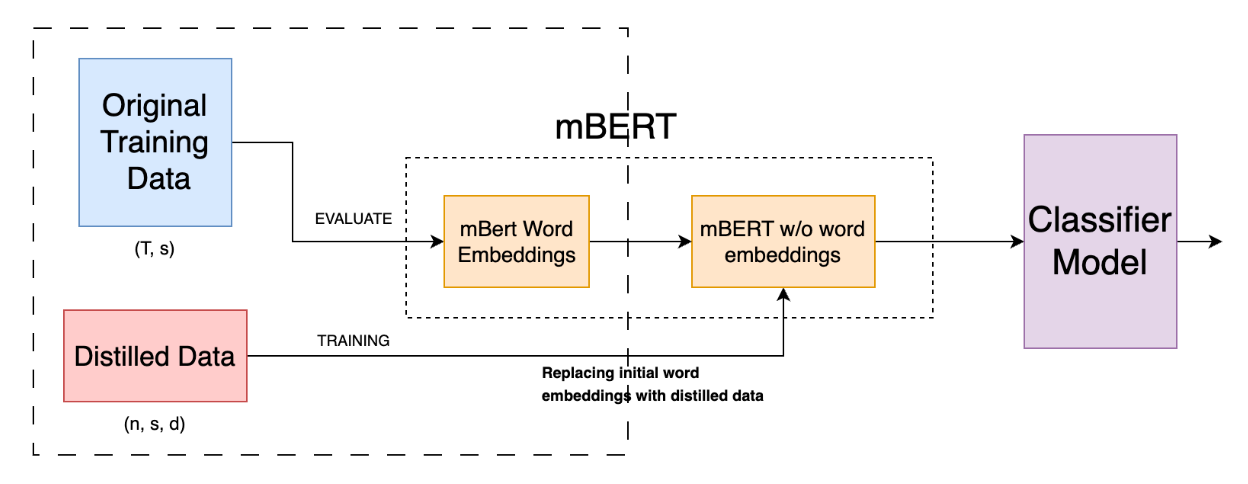}
\caption{SkipLookupDistill Model Architecture: In contrast to Figure \ref{fig:VanillaDistill}, the distilled data now learns at the input level of mBERT, specifically learning mBERT's input word embeddings. Consequently, instead of feeding the distilled data directly to the classifier model, we first pass it to mBERT to get the contextualised embeddings which are then fed to the classifier model.}
\label{fig:SkipLookupDistill}
\end{figure*}
In this method, we take a step back in the distillation process of our VanillaDistill method. We hypothesize that learning at the input level of the language model provides a more robust method for distilling text data. By focusing on learning distilled embeddings at the word-to-embedding level, rather than directly attempting to learn the pre-trained context at the output level of mBERT, we can enhance the interpretability and generalization capability of the distillation process.

Figure \ref{fig:SkipLookupDistill} illustrates the data distillation pipeline for this method, named SkipLookupDistill. Here we utilize the distilled data as word embeddings, bypassing the standard process of mapping word IDs to embedding vectors within the language model's lookup matrix. By utilizing the original lookup matrix, which inherently represents the original dictionary's word embeddings, we can find the most similar words corresponding to our distilled embeddings through nearest neighbor search. This allows us to establish meaningful connections between the distilled embeddings and their corresponding words, enhancing the interpretability and generalization of the distillation process.

\subsection{VocabDistill (Softmax): Distillation at vocab level}

\begin{figure*}[]
\centering
\includegraphics[width=0.7\linewidth]{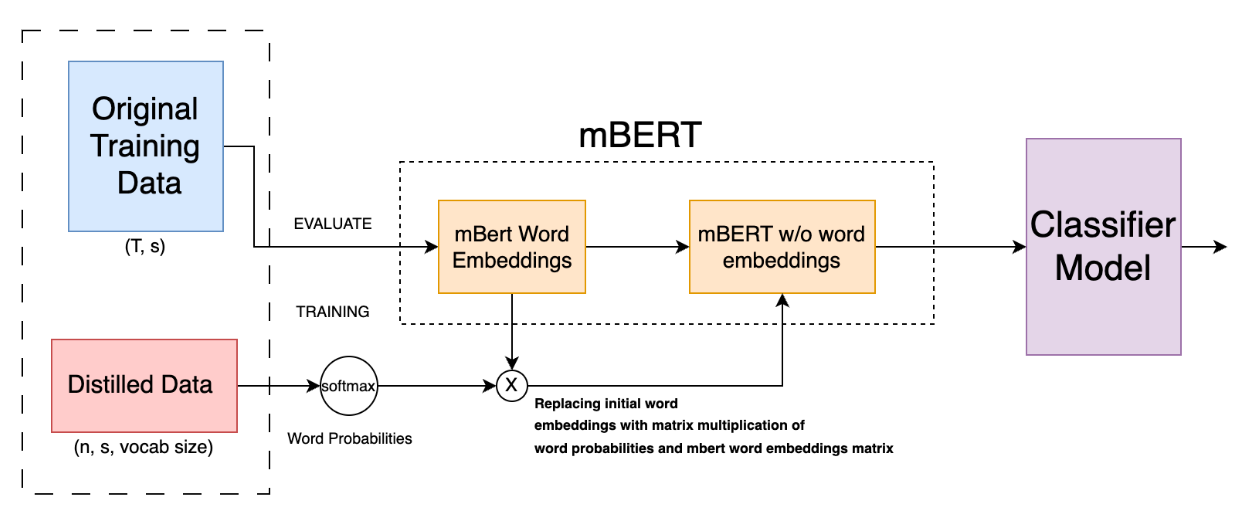}
\caption{VocabDistill (Softmax) Model Architecture: Here we synthesize data at the vocabulary level and apply a Softmax function to obtain word probabilities for each position. These probabilities are then used to create a weighted input word embedding by multiplying them with mBERT's initial word embedding matrix. This approach allows us to incorporate the entire vocabulary distribution rather than a single word. }
\label{fig:vdSoft}
\end{figure*}
In this approach, we employ a refined approach to synthesizing words. Rather than generating embeddings, as we did for the previous methods, we generate word probability vectors of size (number of distilled sentences, sentence length, vocab size). These vectors are then used in conjunction with the softmax function to calculate probabilities for words at each position. We subsequently multiply this probability matrix with mBERT's internal embedding lookup matrix to obtain a weighted embedding input. See Figure \ref{fig:vdSoft} for the architecture diagram. This method aims to enhance word representation generation by incorporating probability-based synthesis and leveraging mBERT's existing embedding lookup matrix. 
With this approach, we gain the advantage of obtaining the distilled words directly. This is achieved by selecting the word with the highest probability for each respective position. As a result, this approach offers increased explainability, as the reasoning behind the selection of words is part of the training process.

\subsection{VocabDistill (Gumbel): Distillation at vocab level}
\begin{figure*}[]
\centering
\includegraphics[width=0.7\linewidth]{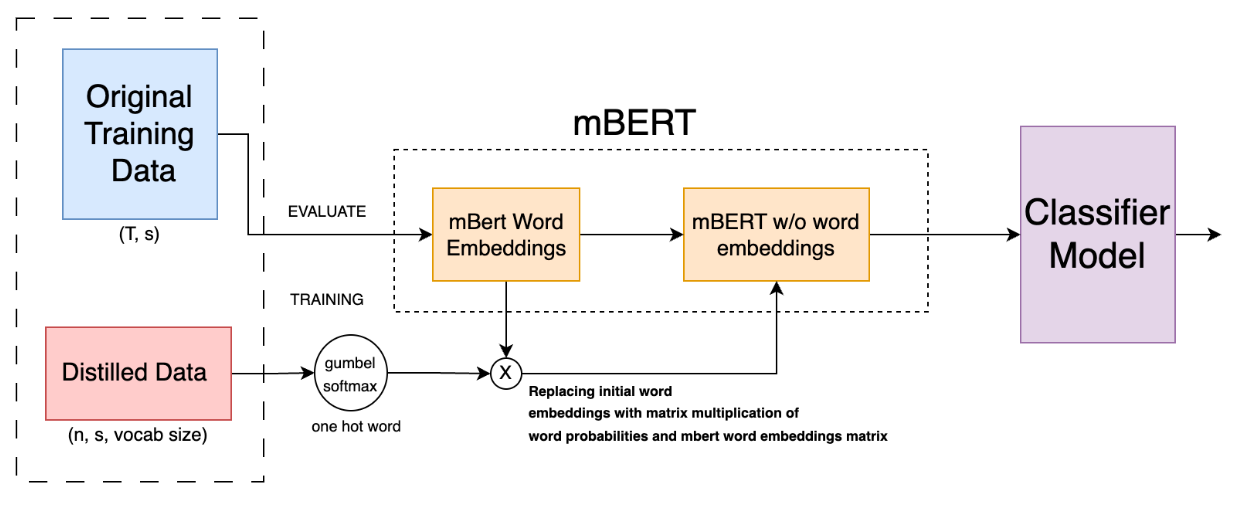}
\caption{VocabDistill (Gumbel): Unlike the model architecture depicted in Figure \ref{fig:vdSoft}, the approach used here involves employing Gumbel-Softmax instead of the traditional Softmax function. This enables us to obtain a single word, as opposed to a probability distribution, which is subsequently multiplied with mBERT's input word embeddings matrix.}
\label{fig:vdGumbel}
\end{figure*}
Here we adopt a similar methodology to the Softmax variant, but with one crucial modification. Instead of using the traditional Softmax function, we employ the Gumbel-Softmax function.  See Figure \ref{fig:vdGumbel} for the architecture diagram. This choice enables us to approximate samples from a categorical distribution while retaining gradient information, unlike the naive argmax approach. As a result, utilizing Gumbel-Softmax allows us to obtain a single word rather than a probability distribution given by Softmax. This aspect facilitates a more intuitive word distillation process, which can yield benefits in terms of explainability. To obtain the distilled words, we adopt a similar approach to the Softmax version, whereby we choose the word with the highest probability for each position.

\begin{table*}[t]
\begin{tabular}{lrrrrrrrr}
\hline
\multicolumn{5}{l|}{\textbf{Training on Original Dataset (F1-Score):}} & \multicolumn{4}{l}{\textbf{57.65}} \\ \hline
\multicolumn{1}{l|}{\textbf{Distilled Samples per class:}} & \multicolumn{4}{c|}{\textbf{1}} & \multicolumn{4}{c}{\textbf{10}} \\ \hline
\multicolumn{1}{r|}{\textbf{Initialization:}} & \multicolumn{2}{c|}{\textbf{Random}} & \multicolumn{2}{c|}{\textbf{Fixed}} & \multicolumn{2}{c|}{\textbf{Random}} & \multicolumn{2}{c}{\textbf{Fixed}} \\ \hline
\multicolumn{1}{r|}{\textbf{Metric:}} & \multicolumn{1}{c|}{\textbf{F1}} & \multicolumn{1}{c|}{\bm{$r_1$}} & \multicolumn{1}{c|}{\textbf{F1}} & \multicolumn{1}{c|}{\bm{$r_1$}} & \multicolumn{1}{c|}{\textbf{F1}} & \multicolumn{1}{c|}{\bm{$r_{10}$}} & \multicolumn{1}{c|}{\textbf{F1}} & \multicolumn{1}{c}{\bm{$r_{10}$}} \\ \hline
VanillaDistill & 53.16 & 92.22 & 54.28 & \textbf{94.15} & 53.24 & 92.35 & 54.84 & \textbf{95.13} \\
SkipLookupDistill & 43.83 & 76.02 & 47.33 & 82.09 & 52.16 & 90.48 & 52.47 & 91.01 \\
VocabDistill (softmax) & 16.91 & 29.33 & 30.57 & 53.04 & 16.92 & 29.34 & 45.59 & 79.07 \\
VocabDistill (gumbel) & 23.37 & 40.54 & 26.86 & 46.59 & 27.78 & 48.18 & 34.88 & 60.51 \\ \hline
\end{tabular}
    \caption{Comparison of our proposed methods with different initialization and the number of distilled samples per class. The F1-Score on the original dataset is presented in the top row. We evaluate two metrics, F1-Score and $r_n$ (n-sample distillation ratio), for all our approaches. The results show that the VanillaDistill method achieves the highest performance. Additionally, we observe improved performance across all models by keeping `Fixed' initialization and 10 distilled samples per class.}
\label{tab:f1score}
\end{table*}

\section{Experimental Settings}

\subsection{Dataset}
\textbf{Unified Multilingual Sentiment Analysis Benchmark, UMSAB} \cite{barbieri-etal-2022-xlm} is a collection of Twitter sentiment analysis datasets for eight different languages with all datasets framed as tweet classification with three labels (positive, negative, and neutral). Languages include Arabic, English, French, German, Hindi, Italian, Portuguese, and Spanish. There are 1838, 323, and 869 tweets in the training, development, and test sets, respectively, for each language.

\subsection{Classifier Model}
\label{classifier}
The classification head in our experiments consists of convolutional layers with multiple filter sizes, max-pooling layers, and a fully connected layer. Specifically, we use three conv2D layers with filter sizes of 3, 4, and 5, respectively. We apply rectified linear unit (ReLU) activation function after each convolutional layer to introduce non-linearity and enhance feature extraction capabilities. Subsequently, we use max-pooling layers to capture the most salient features from each filter. Finally, we use a fully connected layer at the end of the architecture for the classification task.

\subsection{Network Initalization}
The data distillation algorithm is a bi-level optimization involving resetting the parameters of the learning algorithm after each update on the distilled data. Inspired by \cite{Sucholutsky_2021}, we experiment with two initialization methods, one where we reset the parameters with random values and the other where we always reset the parameters to a fixed set of values. We compare the performance of both initialization techniques.

\subsection{Evaluation}
For each method described in Section \ref{sec:methods}, we perform dataset distillation to synthesize 1 and 10 samples per class in our classification task. We evaluate the dataset distillation performance using the following metrics:
\begin{itemize}
    \item \textbf{F1-Score} on models trained using distilled data.
    \item \textbf{n-sample distillation ratio}
    $$r_n=100 \% * \frac{\text { [distillation accuracy] }}{\text { [original accuracy] }},$$ where $n$ is the number of distilled samples.
    This is a relative metric to compare the performance of models trained on distilled data and original data, respectively.
\end{itemize}

We conduct our all experiments on a Linux server with NVIDIA RTX A6000 GPUs. 
\section{Results \& Analysis}

\subsection{Dataset Distillation Performance}
\label{res:DDP}
To assess the performance of our distillation methods, we generate distilled data using them and subsequently train the classification models on the distilled datasets. We then compare the performance of these models trained on the distilled datasets with the model trained on the complete dataset.

Table \ref{tab:f1score} shows the F1 scores and distillation ratios achieved on the test data, highlighting the impact of different initialization methods and the total number of distilled samples per class. Notably, the VanillaDistill approach achieves an impressive distillation ratio of 95\%. This means that a model trained with just 10 samples per class can achieve comparable results to a model trained on the full dataset (consisting of 14K samples). These results clearly demonstrate the potential of text-based dataset distillation. Furthermore, as expected, we observe that as the number of distilled data samples increases, both the F1 score and distillation ratio improve. We also observe that the SkipLookupDistill approach demonstrates comparable performance to VanillaDistill. However, we see that both variants of VocabDistill struggle to achieve good performance. We expect this is due to the large number of parameters in these approaches, which ultimately hinders convergence.

\subsection{Cross-Architecture Generalization}

\begin{table*}[t]
\begin{tabular}{lrrrr}
\hline
\multicolumn{5}{c}{\textbf{Cross Architecture Generalization: F1 Scores on test data}} \\ \hline
\multicolumn{1}{r|}{\textbf{Architecture:}} & \multicolumn{1}{c|}{\textbf{Original}} & \multicolumn{1}{c|}{\textbf{\begin{tabular}[c]{@{}c@{}}Original +\\ 1 FC layer\end{tabular}}} & \multicolumn{1}{c|}{\textbf{\begin{tabular}[c]{@{}c@{}}Original +\\ 2 FC layer\end{tabular}}} & \multicolumn{1}{c}{\textbf{\begin{tabular}[c]{@{}c@{}}Original +\\ 3 FC layer\end{tabular}}} \\ \hline
\multicolumn{1}{l|}{\textbf{Trained on (10 distilled samples per class):}} & \multicolumn{4}{l}{} \\ \hline
VanillaDistill Data & 54.84 & 24.33 & 43.81 & 17.33 \\
SkipLookupDistill Data & 52.47 & \textbf{48.3} & \textbf{50.79} & \textbf{47.15} \\
VocabDistill (softmax) Data & 45.59 & 32.8 & 29.17 & 32.48 \\
VocabDistill (gumbel) Data & 34.88 & 29.94 & 16.69 & 31.94 \\ \hline
\end{tabular}
\caption{Evaluation of the cross-architecture generalization capabilities of our proposed approaches. We generate distilled data using each of the proposed methods and train various model variants using this synthesized data. Subsequently, we assess the trained models on the test set and report their F1-Score. }
\label{tab:crossarc}
\end{table*}

To further evaluate the applicability of our dataset distillation methods, we can use the distilled datasets, generated using the standard classifier model as the learning algorithm (Section \ref{classifier}), to train different architectures. In our experiments, we introduce modifications to the original architecture by using additional fully connected layers. We then evaluate the performance of these classification architectures on the test data.
Table \ref{tab:crossarc} shows the F1 scores of the different model variants when trained on the distilled data generated by our data distillation approaches. The inconsistent performance of the VanillaDistill approach across different architectures suggests that it is sensitive to architectural changes and may not generalize well in all scenarios. Whereas, the SkipLookup method consistently performs well across various architectures, indicating its robustness and the ability to synthesize datasets that are generalizable to different model architectures. This supports our hypotheses (stated in Section \ref{meth:skiplookup}) that learning distilled embeddings at the word-to-embedding level can enhance the generalization capability of the distillation process.
On the other hand, the VocabDistill methods suffer from convergence issues (as observed from the results in Section \ref{res:DDP}). This makes it difficult for us to comment on their generalization capabilities confidently. However, it can be observed that the VocabDistill (softmax) method consistently maintains its performance, albeit at a lower level, across the different model variants. We believe that further experimentation and a deeper understanding of the VocabDistill method can present new avenues for developing more robust and generalizable dataset distillation methods.

\subsection{Language-wise Performance}

\begin{table*}[t]
\begin{tabular}{llrrrr}
\hline
\multicolumn{1}{l|}{} & \multicolumn{5}{c}{\textbf{Training Data (Distilled data has 10 samples per class)}} \\ \hline
\multicolumn{1}{l|}{\textbf{Language}} & \multicolumn{1}{l|}{\textbf{Full Data}} & \multicolumn{1}{c|}{\textbf{VanillaDistill}} & \multicolumn{1}{c|}{\textbf{SkipLookupDistill}} & \multicolumn{1}{c|}{\textbf{\begin{tabular}[c]{@{}c@{}}VocabDistill\\ (softmax)\end{tabular}}} & \multicolumn{1}{c}{\textbf{\begin{tabular}[c]{@{}c@{}}VocabDistill\\ (gumbel)\end{tabular}}} \\ \hline
Arabic & 56.65\% & 49.93\% & \textbf{51.22\%} & 46.57\% & 22.14\% \\
English & 60.71\% & \textbf{62.05\%} & 60.66\% & 50.41\% & 32.83\% \\
French & 63.27\% & \textbf{58.84\%} & 54.11\% & 44.05\% & 40.94\% \\
German & 57.21\% & \textbf{56.91\%} & 53.73\% & 42.41\% & 32.91\% \\
Hindi & 44.63\% & 44.11\% & \textbf{46.48\%} & 41.57\% & 27.38\% \\
Italian & 59.66\% & \textbf{58.14\%} & 52.10\% & 46.67\% & 37.66\% \\
Portuguese & 56.93\% & \textbf{55.85\%} & 50.96\% & 41.27\% & 30.07\% \\
Spanish & 51.70\% & \textbf{51.97\%} & 49.29\% & 39.40\% & 35.32\% \\ \hline
\end{tabular}
\caption{Comparison of language-level performance of our proposed approaches.}
\label{tab:lang}
\end{table*}
The performance of each method across different languages is presented in Table \ref{tab:lang}. We also compare the token-level language proportion between the original data and the distilled data with 10 samples per class, generated using each dataset distillation method. See Figure \ref{fig:langprop} in the Appendix (Section \ref{sec:appendix}). For this, we use an off-the-shelf language detection model: an mBERT model fine-tune for this task. We find that the VanillaDistill approach exhibits a considerably higher percentage of tokens
belonging to languages beyond the original dataset’s eight languages. Furthermore, our remaining three methods
display a similar distribution of token-level languages.

\subsection{Distilled Data Example}
\label{example}
Each distilled embedding can be mapped to its nearest word embedding using the methods described in Section \ref{sec:methods}. Figure \ref{fig:DD} in the Appendix (Section \ref{sec:appendix}) shows the text-mapped distilled data - generated using the VanillaDistill method (best performing). It is worth noting that a model trained on embeddings of just 3 sentences is able to achieve a distillation ratio of 92.22\%.

\section*{Conclusion and Future Work}
In this paper, we address some of the key challenges that hinder the widespread application of dataset distillation methods. Firstly, we explore the application of dataset distillation to text-based datasets. The proposed VanillaDistill and SkipLookupDistill approaches show impressive performance (up to 95\% distillation ratio) on multilingual text classification datasets. Secondly, we extend dataset distillation to multilingual data, which enabled us to perform language-specific analyses on the distilled datasets. Thirdly, we show that the proposed SkipLookupDistill method addresses the challenge of cross-architecture generalization in contrast to the VanillaDistill method. With our experiments, we conclude that this method is more robust, interpretable, and can generate more generalizable distilled datasets. Our VocabDistill methods also show great potential and a deeper understanding of them can present new avenues for developing more generalizable dataset distillation methods. To the best of our knowledge, our work is the first to address cross-architecture generalization in text-based dataset distillation. With this work, we take a step towards filling the following gaps in the ongoing research - cross-architecture generalization, improving interpretability, and leveraging context in text dataset distillation.

For future work, it is worth experimenting with a broader range of datasets and better classification models including fine-tuning language models. This will allow for a more comprehensive evaluation of the proposed methods. Furthermore, it would be interesting to use learnable labels instead of fixed labels, which we currently use in our experiments, to further enhance the data distillation process.

\section*{Acknowledgements}
We are thankful to Prof. Jingo Shang, Bill Willam Hogan, and Noveen Sachdeva for their constant support and guidance throughout this work.

\bibliography{custom}
\bibliographystyle{acl_natbib}

\appendix
\section{Appendix}
\label{sec:appendix}
\begin{figure}[h]
\centering
\includegraphics[width=\linewidth]{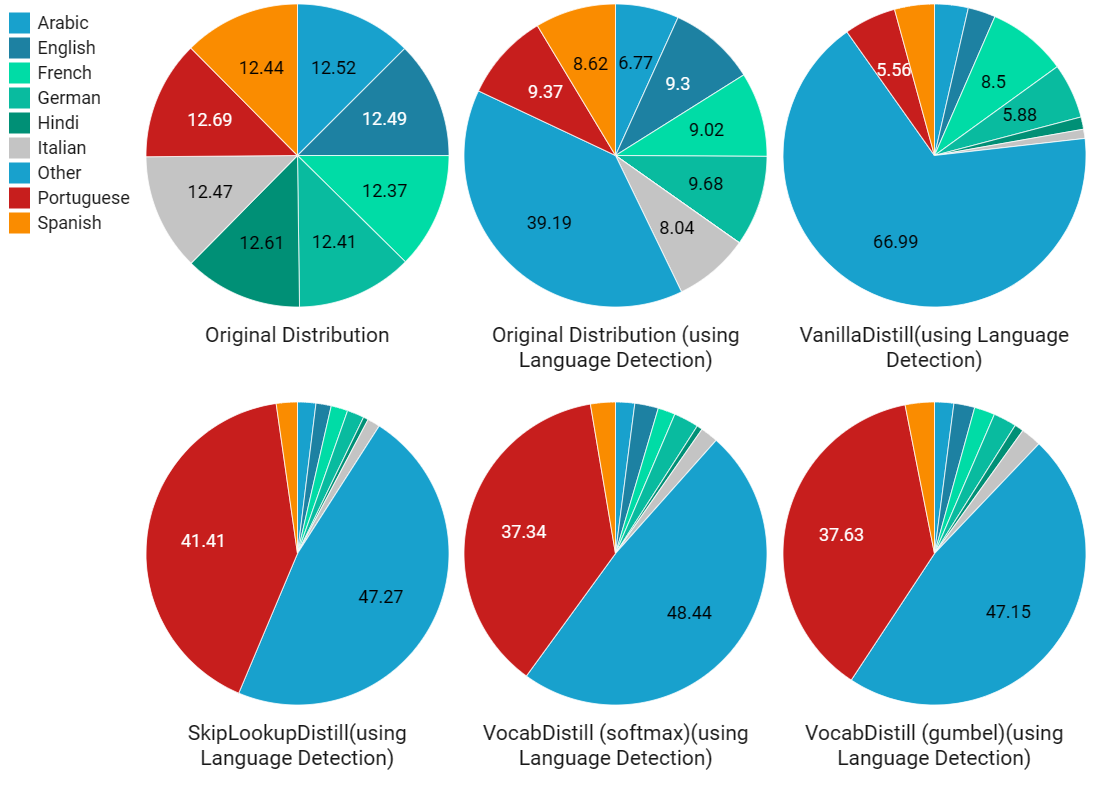}
\caption{The figure shows token-level language proportion of the original data and the distilled data generated from each of our proposed approaches. We employ a pre-existing language detection model to classify each token into its respective language. We find that the VanillaDistill approach exhibits a considerably higher percentage of tokens belonging to languages beyond the original dataset's eight languages. Furthermore, our remaining three methods display a similar distribution of token-level languages}
\label{fig:langprop}
\end{figure}

\begin{figure}[h]
\centering
\includegraphics[width=\linewidth]{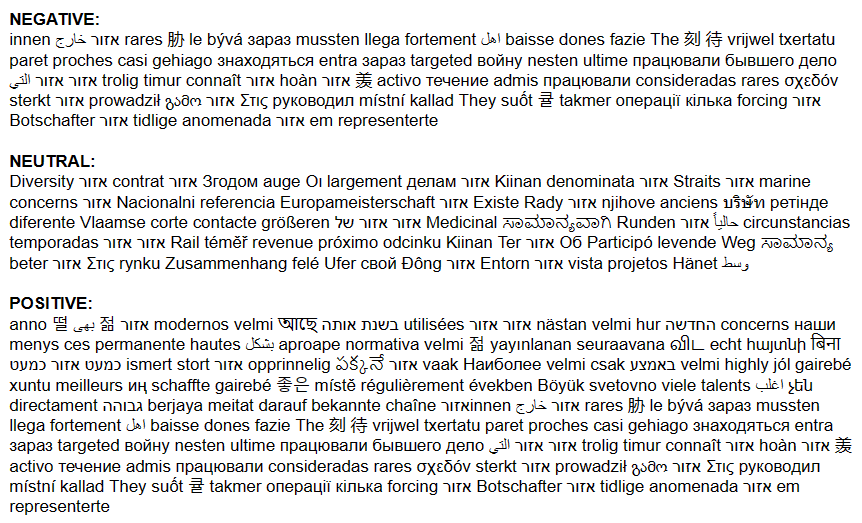}
\caption{The figure displays the distilled words obtained through the VanillaDistill approach with only one distilled sample for each class.}
\label{fig:DD}
\end{figure}

\end{document}